\def\year{2021}\relax
\documentclass[letterpaper]{article} 
\usepackage{aaai21}  
\usepackage{times}  
\usepackage{helvet} 
\usepackage{courier}  
\usepackage[hyphens]{url}  
\usepackage{graphicx} 
\usepackage{amstext}
\usepackage{natbib}  
\usepackage{caption} 
\usepackage{multirow}
\usepackage{colortbl,booktabs}
\usepackage{color}
\usepackage[switch]{lineno}  %
{\color{red}}

\title{DialogXL: All-in-One XLNet for Multi-Party Conversation Emotion Recognition}

\author{
    Weizhou Shen,
    Junqing Chen,
    Xiaojun Quan\thanks{Corresponding author.},
    Zhixian Xie
    \\
}

\affiliations{

    Sun Yat-sen University, China\\
    \{shenwzh3, chenjq95, xiezhx23\}@mail2.sysu.edu.cn, quanxj3@mail.sysu.edu.cn

}

\begin{document}
\maketitle

\begin{abstract}
	This paper presents our pioneering effort for emotion recognition in conversation (ERC) with pre-trained language models. Unlike regular documents, conversational utterances appear alternately from different parties and are usually organized as hierarchical structures in previous work. Such structures are not conducive to the application of pre-trained language models such as XLNet. To address this issue, we propose an all-in-one XLNet model, namely DialogXL, with enhanced memory to store longer historical context and dialog-aware self-attention to deal with the multi-party structures. Specifically, we first modify the recurrence mechanism of XLNet from segment-level to utterance-level in order to better model the conversational data. Second, we introduce dialog-aware self-attention in replacement of the vanilla self-attention in XLNet to capture useful intra- and inter-speaker dependencies. Extensive experiments are conducted on four ERC benchmarks with mainstream models presented for comparison. The experimental results show that the proposed model outperforms the baselines on all the datasets. Several other experiments such as ablation study and error analysis are also conducted and the results confirm the role of the critical modules of DialogXL.
\end{abstract}

\section{Introduction}
Emotion recognition in conversation (ERC) is an emerging task in natural language processing (NLP) that aims to identify the emotion of each utterance in a conversation. It can be regarded as an extension of traditional emotion detection from text, or an arising problem in dialogue systems that helps generate emotion-aware dialogues \cite{zhou2017emotional}. Empirical evidence shows that the conversational context of an utterance plays an indispensable role in this task \cite{poria2019emotion}. Moreover, the emotion also tends to stay unchanged within a short context of the conversation. It is thus very critical to effectively model the alternate utterances by different parties.

To solve this problem, many recent works focus on deep neural networks with hierarchical structures to model the conversational data \cite{majumder2019dialoguernn, ghosal2019dialoguegcn,jiao2019higru,zhong2019knowledge}. In these works, each utterance is firstly encoded separately into an utterance representation, which is then modeled sequentially and hierarchically. Although the structures seem to comply with the organization of utterances, they ignore the direct dependencies between words in different utterances. In addition, they are not conducive to the application of pre-trained language models such as BERT \cite{devlin2018bert} and XLNet \cite{yang2019xlnet}, which have achieved superior performance in many dialogue system tasks other than ERC \cite{madotto2018mem2seq,bao2020plato,henderson2019convert}. 

There are two main challenges to directly apply these pre-trained language models to ERC. First, conversations in ERC are usually multi-party and there can be intra- and inter-speaker dependencies \cite{ghosal2019dialoguegcn}. Existing pre-trained language models are not readily feasible to encode these dependencies. Second, almost all language models are constrained by the input length. When the input sequence exceeds the limit, it has to be truncated, which may lead to loss of information in distant historical utterances \cite{majumder2019dialoguernn,ghosal2019dialoguegcn}.

To cope with the above challenges, we introduce an all-in-one XLNet model, namely DialogXL, for emotion recognition in multi-turn multi-party conversation. DialogXL intends to apply a strong pre-trained language model to ERC without constructing a complicated, hierarchical model in processing the conversational data. Specifically, it first replaces XLNet's \emph{segment recurrence} by a more flexible and memory-saving \emph{utterance recurrence} to utilize historical utterances. Utterance recurrence stores the hidden states of historical utterances in a memory bank and reuses them while identifying a query utterance. Next, the \emph{self-attention} in XLNet's Transformer layers is substituted for \emph{dialog-aware self-attention}, which consists of four different types of attention, namely local self-attention, global self-attention, speaker self-attention, and listener self-attention. Dialog-aware self-attention allows DialogXL to model the inter- and intra-speaker dependencies under different reception fields in the historical context. We conduct extensive experiments on four ERC benchmarks and the results show that the proposed model, DialogXL, outperforms all the baselines on the datasets. Furthermore, several studies are conducted to verify the modules of DialogXL, and an error analysis is used to delve into the reasons behind the errors.

To conclude, our contributions are as follows:
\begin{itemize}
\item DialogXL is the first effort of pre-trained language models designed for emotion recognition in conversation (ERC).

\item We propose a memory-saving utterance recurrence to replace XLNet's segment recurrence. The new approach allows DialogXL to cache up to 1000 historical words of a conversation, which is more powerful than the vanilla XLNet model.

\item  Unlike the original self-attention that merely computes attention weights between words, our dialog-aware self-attention computes them by different reception fields and party roles, allowing us to capture  useful intra- and inter-speaker dependencies.
\end{itemize}

\section{Related Work}
\subsection{Emotion Recognition in Conversation}
Emotion recognition in conversation (ERC) has emerged as an important problem in recent years and has attracted numerous interests from the NLP community. The availability of large conversational datasets \cite{busso2008iemocap, schuller2012avec, li2017dailydialog, chen2018emotionlines, poria2019meld} account partly for this phenomenon, and the increasing interests in dialogue systems may also explain it. 

Recent works on ERC generally resort to deep learning models. For example, CMN \cite{hazarika2018conversational} and ICON \cite{hazarika2018icon} both utilize gated recurrent unit (GRU) and memory networks. \citeauthor{majumder2019dialoguernn} \shortcite{majumder2019dialoguernn} propose a recurrent-based model to model the party state, global state and emotional dynamics. \citeauthor{jiao2019higru} \shortcite{jiao2019higru} propose a hierarchical GRU structure that trains utterance-level and conversation-level encoders jointly. \citeauthor{ghosal2019dialoguegcn} \shortcite{ghosal2019dialoguegcn} propose a graph neural network based model to encode speaker dependencies and temporal information. \citeauthor{zhong2019knowledge}  \shortcite{zhong2019knowledge} incorporate external knowledge bases to support the identification. \citeauthor{hazarika2019emotion} \shortcite{hazarika2019emotion} introduce transfer learning from utterance generation to ERC.

The modalities of data used in the above works are not the same. Specifically, \cite{hazarika2018icon,hazarika2018conversational,majumder2019dialoguernn} utilize textual, audio and video modalities, while the latest research \cite{jiao2019higru,ghosal2019dialoguegcn,zhong2019knowledge,hazarika2019emotion} tends to use only the textual modality.

\subsection{Pre-trained Language Models}
The effectiveness of large pre-trained language models \cite{devlin2018bert, yang2019xlnet, liu2019roberta, conneau2020unsupervised} has been well exhibited in many NLP tasks such as machine reading comprehension, text classification, machine translation. Among the language models, BERT \cite{devlin2018bert} utilizes bi-directional Transformer encoders as well as pre-training schemes of masked language modeling and next sentence prediction. XLNet \cite{yang2019xlnet} is another powerful pre-trained language model, which excels at processing long documents with the segment recurrence mechanism. In addition, it combines the strengths of both auto-encoding and auto-regressive language modeling. There have been some recent works that apply pre-trained language models to  dialog-related tasks \cite{bao2020plato,ham2020end, henderson2019convert}, but they have yet to be applied to emotion recognition in conversation.

\begin{figure*}[t]
	\centering
	\includegraphics[width=1\textwidth]{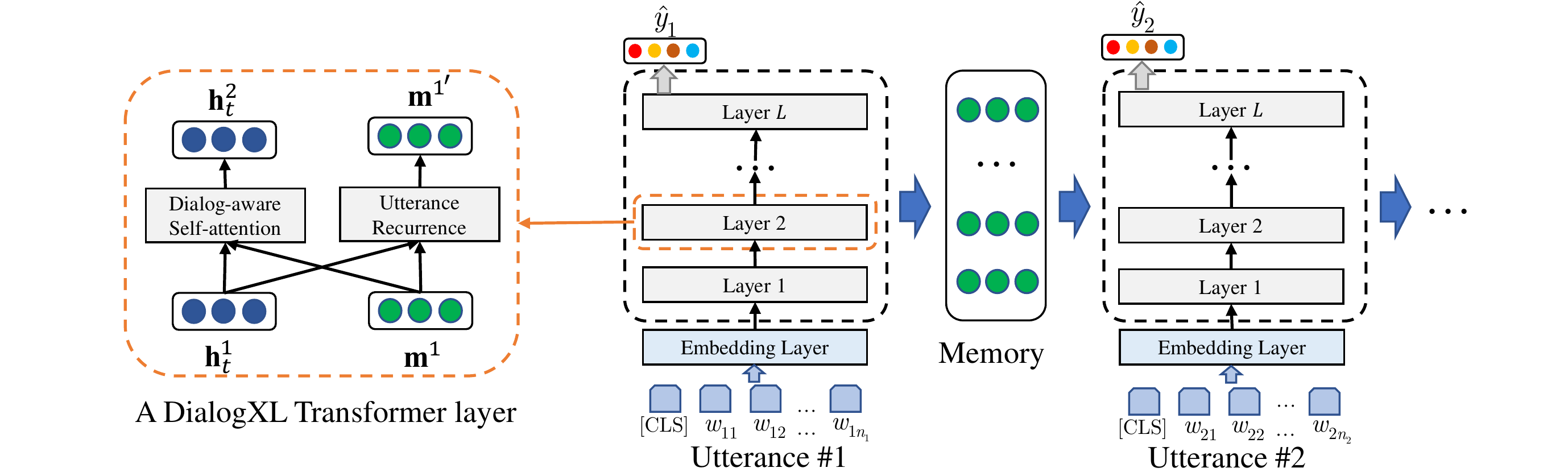} 
	\caption{The architecture of our DialogXL.}
	\label{fig:model_overview}
\end{figure*} 

\section{Methodology}
There are two challenges to overcome in order to apply pre-trained language models to emotion recognition in conversation (ERC). The first challenge is how to encode a long historical context with hundreds of words. The second is how to model the intra- and inter-speaker dependencies of different parties. Instead of building a hierarchical network as previous work, we propose DialogXL\footnote{The implementation is available at https://github.com/shen-wzh3/DialogXL.} to address these two challenges on the basis of XLNet with two improvements. 

The overview architecture of DialogXL is shown in Figure \ref{fig:model_overview}. It consists of an embedding layer, 12 Transformer layers, and a feed-forward neural network. The model identifies the emotion for each utterance in turn when a conversation comes in. Compared with XLNet, DialogXL has a more effective memory bank equipped during the training and testing phases, storing hidden states of historical utterances for future reuses. The memory bank is updated by a new \emph{utterance recurrence} mechanism. And the hidden states at each Transformer layer are derived by \emph{dialog-aware self-attention}.\footnote{XLNet's permutation language modeling and two-stream self-attention, which are designed for language modeling tasks, are not included in DialogXL.}

\subsection{Problem Definition}
In ERC, a conversation is defined as a list of utterances $\{u_1, u_2, ..., u_N\}$, where $N$ is the number of utterances. Each utterance $u_i$ consists of $n_i$ tokens, namely $u_i = \{w_{i1}, w_{i2},...,w_{in_i}\}$. A discrete value $y_i\in \mathcal{S}$ is used to denote the emotion label of $u_i$, where $\mathcal{S}$ is the set of emotion labels. The speaker is denoted by a function $p(\cdot)$. For example, $p(u_i)\in \mathcal{P}$ denotes the speaker of $u_i$ and $\mathcal{P}$ is the collection of all speaker roles in an ERC dataset. The objective of this task is to output the emotion label $y_t$ for a given query utterance $u_t$ based on its historical context $\{u_1, u_2, ..., u_{t-1}\}$ and the corresponding speaker information.

\subsection{Model Input}
At each time step $t$, the query utterance $u_t$ is prepended with the special token ``\texttt{[CLS]}'':
\begin{equation}
x_t = \{\text{[CLS]}, w_{t1}, w_{t2}, ..., w_{tn_t}\}.
\end{equation}
The utterance is then passed to the embedding layer. In DialogXL, this layer consists of only word embedding. The output of the embedding layer is treated as input hidden states to the first Transformer layer:
\begin{equation}
\mathbf{h}_t^0 = \text{Embedding}(x_t)
\end{equation}

\subsection{Utterance Recurrence}

XLNet \cite{yang2019xlnet} and Transformer-XL \cite{dai2019transformer} address the limitation of input size by a mechanism named \emph{segment recurrence}, which caches previous hidden states in a memory bank and revisits them in future computations. However, this mechanism is ineffective when directly applied to conversational emotion recognition for two reasons. First, the ``segment'' in XLNet refers to a fixed-length sequence rather than a linguistic unit such as a sentence. The conversation in ERC is defined in terms of utterances, which are typically full sentences or paragraphs. Therefore, it is essential to keep the utterances complete rather than segmented into pieces. Second, segment recurrence constrains segments in the same training batch to have the same length, which results in too many paddings stored in memory. By contrast, the proposed \emph{utterance recurrence} stores the historical context in memory without paddings, allowing the memory to store a longer historical context. 

\begin{figure}[t]
	\centering
	\includegraphics[width=1\columnwidth]{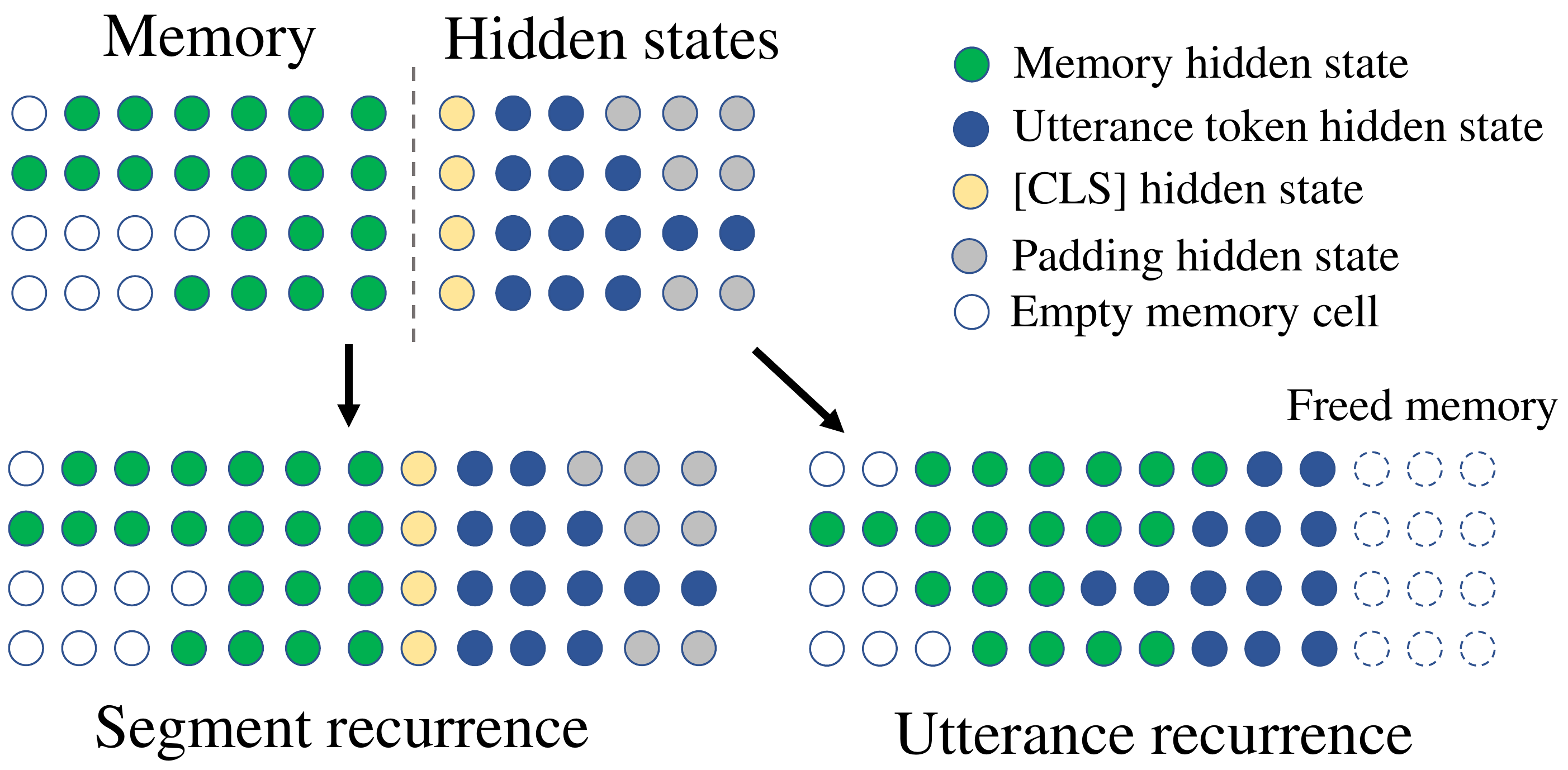} 
	\caption{Illustration of the memory update strategies by utterance recurrence and segment recurrence. The batch size is 4, with each row corresponding to a conversation.}
	\label{fig:utterance_recurrence}
\end{figure}

\begin{figure*}[t]
	\centering
	\includegraphics[width=0.9\textwidth]{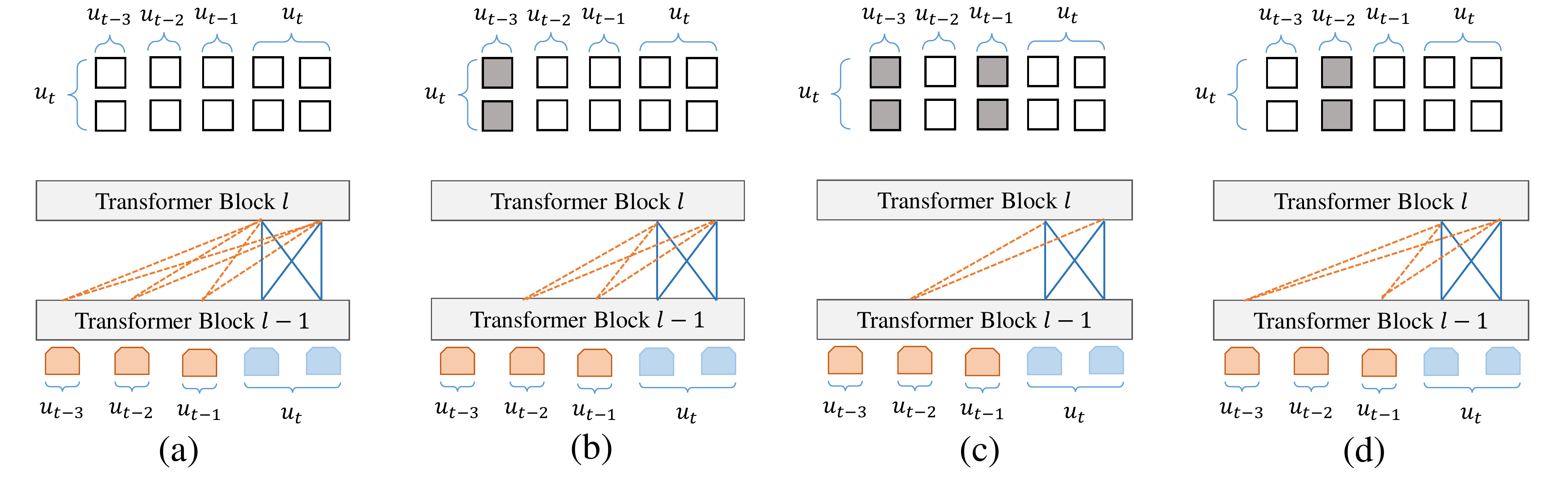} 
	\caption{Demonstration of dialog-aware self-attention: (a) global self-attention, (b) local self-attention, (c) speaker self-attention, and (d) listener self-attention. The utterance to be identified is $u_t$, while the hidden states of $u_{t-3}$, $u_{t-2}$, and $u_{t-1}$ are cached in memory. The speaker identities are: $p(u_t)=p(u_{t-2}) $ and $p(u_{t-1})=p(u_{t-3})$. The window size of local self-attention is 2. The upper part of each subgraph is the attention mask for $u_t$ and other utterances, with masked attention weights colored grey. The lower part of each subgraph is the attention flows between two consecutive Transformer layers, where solid blue lines represent self-attention within $u_t$ and dashed orange lines represent attention between $u_t$ and the memory $\mathbf{m}$.}
	\label{fig:multi_perspective_attention}
\end{figure*} 

The memory, denoted by $\mathbf{m}$, works like a stack. Every time a new set of hidden states are generated for a query utterance, they are concatenated with the current memory. To prevent from introducing noises into the memory, only the hidden states of the utterance tokens are stored, with the hidden states of the ``\texttt{[CLS]}'' and padding positions ignored. Formally, for the $t$-th utterance, at each Transformer layer $l$ the new memory $\mathbf{m}^{l'}$ is updated as:
\begin{equation}
\mathbf{m}^{l'} = \mathbf{m}^{l} \parallel \mathbf{h}_{t,1:1+n_t}^{l}
\end{equation} 
where $\parallel$ denotes the concatenation operation. This update strategy is useful especially during the batching operation. As illustrated in Figure \ref{fig:utterance_recurrence}, updating memory with only the hidden states of utterance tokens makes the memory more compact, for the noises introduced by padding are mostly eliminated and more space is freed to cache a longer context.

\subsection{Dialog-Aware Self-Attention}
Utterances occur alternately by different parties in a conversation, and the vanilla self-attention in XLNet cannot be directly applied to the  multi-party setting. To this end, we replace the self-attention by \emph{dialog-aware self-attention}, which enables our model to encode conversation contexts in a multi-turn multi-party setting. The new self-attention consists of four types of self-attention: \emph{global self-attention} and \emph{local self-attention} for different sizes of receptive fields, and \emph{speaker self-attention} and \emph{listener self-attention} for intra- and inter-speaker dependencies. We implement the dialog-aware self-attention by skillfully changing the masking strategies of self-attention, without the need to add any extra embeddings or parameters, as illustrated in Figure \ref{fig:multi_perspective_attention}.

Dialog-aware self-attention is multi-headed. For each attention head of the $l$-th Transformer layer, the attention output $\mathbf{o}^l_t$ is computed as follows:
\begin{equation}\label{eq:attn_begin}
\widetilde{\mathbf{h}}_t^{l-1} = \mathbf{m}^{l-1} || \mathbf{h}_t^{l-1}
\end{equation} 
\begin{equation}
\mathbf{q}^l_t,\ \mathbf{k}^l_t,\ \mathbf{v}^l_t = \mathbf{h}_t^{l-1}\mathbf{W}^{l\top}_q,\ \widetilde{\mathbf{h}}_t^{l-1}\mathbf{W}^{l\top}_k,\ \widetilde{\mathbf{h}}_t^{l-1}\mathbf{W}^{l\top}_v
\end{equation}
\begin{equation}
\mathbf{a}_t^l = \text{RelPosAttn}(\mathbf{q}^l_t, \mathbf{k}^l_t)
\end{equation}
\begin{equation}\label{eq:mask}
\widetilde{\mathbf{a}}_t^l = \mathbf{a}_t^l- \mathbf{s}
\end{equation}
\begin{equation}\label{eq:attn_end}
\mathbf{o}_t^l = \text{softmax}(\widetilde{\mathbf{a}}_t^l) \mathbf{v}^l_t
\end{equation}
where $\mathbf{W}^{l}_q$, $\mathbf{W}^{l}_k$, and $\mathbf{W}^{l}_v$ are trainable parameters for each attention head, and $\text{RelPosAttn}(\cdot)$ are the relative position attention adopted from Transformer-XL and XLNet. 

The attention mask $\mathbf{s}$ in Equation (\ref{eq:mask}) is a matrix with the same shape as the attention weights $\mathbf{a}_t^l$. The value of $\mathbf{s}_{ij}$ is set to $+\infty$ only when the attention between the $i$-th vector in $\mathbf{q}^l_t$ and $j$-th vector in $\mathbf{k}^l_t$ is masked, and set to 0 otherwise.

For the sake of convenience, we denote Equation (\ref{eq:attn_begin}) to Equation (\ref{eq:attn_end}) by a function $f(\cdot)$:
\begin{equation}
	\mathbf{o}_t^l = f(\mathbf{m}^{l-1},\mathbf{h}_t^{l-1},\mathbf{s})
\end{equation}

\subsubsection{Global Self-Attention}
Global self-attention takes all the historical context and the query utterance as the reception field. It is the same as the vanilla self-attention, in which the query utterance pays attention to the whole context. This setting allows our model to attend to previously distant utterances which may also be useful \cite{majumder2019dialoguernn}. Thus, no masking is made for global self-attention:
\begin{equation}
\mathbf{s}^{global}_{ij}=0
\end{equation}

\subsubsection{Local Self-Attention}
Local self-attention only has a reception field of $\omega$ latest historical utterances, where $\omega$ is a hyperparameter. The motivation for this attention is that intuitively speaker's emotion is mostly influenced by the recent utterances. In local self-attention, we mask the attentions between the query utterance and the historical utterances outside the reception field:
\begin{equation}
	\mathbf{s}^{local}_{ij} = \left\{
	\begin{array}{rl}
	+\infty,    &  j\notin \text{Idx}(\{u_{t-\omega},u_{t-\omega+1},...,u_{t-1},u_t\})\\
	0,       & \text{Otherwise}\\
	\end{array} \right.
\end{equation}
where $\text{Idx}(\mathcal{U})$ is a function that maps the utterance tokens in $\mathcal{U}$ to the corresponding positions in the key matrix $\mathbf{k}^l_t$ .

\subsubsection{Speaker Self-Attention}
Speaker self-attention considers only the historical context spoken by the present speaker. It intends to model the intra-speaker dependency \cite{ghosal2019dialoguegcn} by identifying emotional clues in the speaker's historical utterances. In speaker self-attention, we mask the attentions between the query utterance and the utterances spoken by other speakers:
\begin{equation}
\mathbf{s}^{speaker}_{ij} = \left\{
\begin{array}{rcl}
+\infty,  &j\in \text{Idx}(\{u\ |\ p(u)\neq p(u_t)\})\\
0,     &\text{Otherwise}\\
\end{array} \right.
\end{equation}

\subsubsection{Listener Self-Attention}
Listener self-attention considers only the historical utterances spoken by other speakers. It intends to model the inter-speaker dependency \cite{ghosal2019dialoguegcn}, meaning that the present speaker's emotion may be influence by other speakers' words. In listener self-attention, we mask the attentions between the query utterance and the utterances made by the present speaker:
\begin{equation}
\mathbf{s}^{listener}_{ij} = \left\{
\begin{array}{rl}
+\infty, &j\in \text{Idx}(\{u\ |\ p(u)=p(u_t)\})\\
0,  &\text{Otherwise}\\
\end{array} \right.
\end{equation}

The outputs of the four types of self-attention are concatenated and passed through a normalization layer followed by a feed-forward network to generate the output for this Transformer layer:
\begin{equation}
\widetilde{\mathbf{o}}^l_t = \parallel_{k=1}^{K}f_k(\mathbf{m}^{l-1},\mathbf{h}^{l-1}_t,\mathbf{s}^{c_k} )
\end{equation}
\begin{equation}
\mathbf{h}^l_t = \text{FeedForward}(\text{LayerNorm}(\widetilde{\mathbf{o}}^l_t))
\end{equation}
where $K$ is the number of self-attention heads, and $c_k \in\{global,\ local,\  speaker,\ listener\}$ is the corresponding type of dialog-aware attention for the $k$-th attention head.

\subsection{Model Training}
We take the hidden state of ``\texttt{[CLS]}'' at the last layer as the final encoding of the query utterance and the historical context, and pass it through a feed-forward neural network to get the predicted emotion:
\begin{equation}
\mathbf{h}_t = \mathbf{h}^L_{t,0}
\end{equation}
\begin{equation}
\mathbf{z}_t = \text{ReLU}(\mathbf{W}_h\mathbf{h}_t+\mathbf{b}_h)
\end{equation}
\begin{equation}
P_t = \text{softmax}(\mathbf{W}_z\mathbf{z}_t+\mathbf{b}_z)
\end{equation}
\begin{equation}
\widehat{y}_t = \text{argmax}_{k\in\mathcal{S}}(P_t[k])
\end{equation}

For the training of our model, we use the standard cross-entropy loss as the loss function:
\begin{equation}
\mathcal{L}(\theta) = -\sum_{i=1}^{M}\sum_{t=1}^{N}P_t[y_{i,t}]
\end{equation}
where $M$ is the number of conversations in the training set, and $\theta$ is the collection of trainable parameters in DialogXL.

\section{Experimental Settings}

In this section, we present the experimental settings such as implementation details, datasets, metrics, and baselines.

\subsection{Implementation Details}
We initialize the proposed DialogXL by pre-trained XLNet-Base \cite{yang2019xlnet} and employ AdamW optimizer \cite{loshchilov2018fixing} during training. Hyperparameter tuning for each dataset is conducted  with hold-out validation on the validation set. The tunable hyperparameters include learning rate, number of heads for the four types of attentions in dialog-aware self-attention\footnote{The sum of the four types of attention heads is always 12.}, the max length of memory\footnote{When implementing utterance recurrence in practice, the length of memory is set not to exceed a threshold due to the limit of computational resource. Once the size of memory exceeds the threshold, the earliest hidden states will be dropped. }, and the dropout rate. The results of BERT, XLNet, and DialogXL reported in our experiments are all based on the average score of 5 random runs on the test set.

\subsection{Datasets}
We evaluate DialogXL on four multi-turn multi-party ERC datasets. The statistics of them are shown in Table \ref{tab:statistic}.
\begin{table}[t]
	\centering
	\resizebox{0.475\textwidth}{!}{
	\begin{tabular}{l|c|c|c|c|c|c}
		\hline
		\multirow{2}*{Dataset} & \multicolumn{3}{|c|}{\# Conversations} &\multicolumn{3}{|c}{\# Uterrances}\\ 
		&Train&Val&Test&Train&Val&Test\\
		\hline
		\hline
		IEMOCAP&\multicolumn{2}{c|}{120}&31&\multicolumn{2}{c|}{5810}&1623\\ 
		MELD&1038&114&280&9989&1109&2610\\
		DailyDialog&11118&1000&1000&87170&8069&7740\\
		EmoryNLP&713&99&85&9934&1344&1328\\
		\hline
	\end{tabular}
	}
	\caption{The statistics of four datasets.}
	\label{tab:statistic}
\end{table}

\noindent\textbf{IEMOCAP} \cite{busso2008iemocap}: A multimodal  conversational dataset for emotion recognition, with two parties included for each conversation. The emotion labels include \texttt{neutral}, \texttt{happiness}, \texttt{sadness}, \texttt{anger}, \texttt{frustrated}, and \texttt{excited}. Since this dataset has no validation set, we follow \cite{zhong2019knowledge} to use the last 20 dialogues in the training set for validation.

\noindent\textbf{MELD} \cite{poria2019meld}: A multimodal dataset for emotion recognition collected from the TV show \texttt{Friends}. The emotion labels include \texttt{neutral}, \texttt{happiness}, \texttt{surprise}, \texttt{sadness}, \texttt{anger}, \texttt{disgust}, and \texttt{fear}. 

\noindent\textbf{DailyDialog} \cite{li2017dailydialog}: Human-written daily communications, with emotion labels including \texttt{neutral},   \texttt{happiness}, \texttt{surprise}, \texttt{sadness}, \texttt{anger}, \texttt{disgust}, and \texttt{fear}. Since it has no speaker information, we consider the utterance turns as the speaker turns by default.

\noindent\textbf{EmoryNLP} \cite{zahiri2017emotion}: TV show scripts collected from \texttt{Friends}, but varies from MELD in the choice of scenes and emotion labels. The emotion labels of this dataset include \texttt{neutral}, \texttt{sad}, \texttt{mad}, \texttt{scared}, \texttt{powerful}, \texttt{peaceful}, and \texttt{joyful}.

Following recent works \cite{ghosal2019dialoguegcn,zhong2019knowledge,hazarika2019emotion}, we utilize only the textual data of the above datasets for our experiments. The evaluation metrics are chosen as micro-F1 for DailyDialog\footnote{The category of neutral with significantly more samples than others are not taken into account as before.} and weighted-F1 for the other datasets.

\subsection{Baseline Methods}

We compare DialogXL with the following baselines:

\noindent\textbf{Previous methods:} CMN\cite{hazarika2018conversational}, DialogueRNN \cite{majumder2019dialoguernn}, HiGRU\cite{jiao2019higru}, DialogueGCN\cite{ghosal2019dialoguegcn}, TL-ERC\cite{hazarika2019emotion},  
and KET \cite{zhong2019knowledge}. 

\noindent\textbf{BERT} \cite{devlin2018bert}: The BERT baseline for ERC, initialized with the pre-trained parameters of BERT-base. We concatenate historical utterances and the query utterance in order and then feed them into BERT for classification. The hyperparameters are tuned the same as DialogXL. 

\noindent\textbf{XLNet} \cite{yang2019xlnet}: The XLNet baseline with the original segment recurrence and vanilla self-attention, initialized with the pre-trained parameters of XLNet-base. The hyperparameters are tuned the same as DialogXL.

\section{Results and Analysis}

\subsection{Overall Results}
The overall results of our DialogXL and the baselines are reported in Table \ref{tab:overall}. We can clearly note that DialogXL reaches a new state of the art on all of the four datasets. Besides, we can make another two observations as follows, which help to understand the ERC task and the pros and cons of DialogXL.

First,  in general, there are considerable improvements for the pre-trained language models over the others on MELD, DailyDialog, and EmoryNLP. However, the improvements of DialogXL over BERT and XLNet are not significant on these datasets. After delving into the datasets, we found that the dialogues in these datasets are relatively short (mostly 5 to 9 utterances). So the current language models, BERT and XLNet, can already encode the entire historical context and the query utterance in most cases. On these short dialogues, however, the advantages of DialogXL are not shown up completely. 

Second, while inferior performance of BERT and XLNet is observed to the other baselines on IEMOCAP, the improvements of DialogXL over BERT and XLNet are significant. After examining the dataset, we realized the dialogues in IEMOCAP are much longer (around 70 utterances per dialog) than the other datasets. In this case, BERT and XLNet cannot encode too much historical context effectively, while such baselines as DialogueRNN and DialogueGCN can reach distant utterances and also encode other key features such as speaker information. Moreover, our DialogXL can both encode the historical context effectively by the utterance recurrence and 
capture the speaker information by the dialog-aware self-attention, allowing it to achieve superior performance to all the baselines.

\begin{table}[t]
	\centering
		\resizebox{0.475\textwidth}{!}{
	\begin{tabular}{l|cccc} 
		\hline
		Model & IEMOCAP &MELD &DailyDialog &EmoryNLP\\ 
		\hline
	   CMN&    56.13    &-    &-           &-\\
	   DialogueRNN&62.75&57.03&-           &-\\
		HiGRU&59.79*     
						& 56.92*  &52.01*           &31.88*\\
	   DialogueGCN&64.18&58.10&-           &-\\
		TL-ERC&59.30     &57.46*    &52.46*       &30.57*\\
		KET&59.56       &58.18&53.37       &33.95*\\
		BERT&60.98      &61.50&54.09       &34.17\\
		XLNet&61.33         &61.65&53.62       &34.13\\
		\hline
		DialogXL &\textbf{65.94}&\textbf{62.41}&\textbf{54.93}&\textbf{34.73}\\
		\hline
	\end{tabular}
	}
	\caption{Overall performance on the four datasets. The scores marked by ``*'' is based on our re-implementation, because of the differences in evaluation metrics and data statistics between the corresponding work and ours.}
	\label{tab:overall}
\end{table}

\subsection{Effect of the Enhanced Memory}
One of the contributions of DialogXL lies in the enhanced memory with utterance recurrence. Here, we study how the utterance recurrence and the maximum memory length contribute to the final results. We change the maximum memory length from 100 to 1000 with an interval of 100 and plot the test scores on IEMOCAP, which has sufficient utterances in each conversation. The memory waste rate of segment recurrence in XLNet is also plotted in terms of the percentage of paddings in memory. Since the proposed utterance recurrence has 0 memory waste in theory, its memory waste rate is not plotted. Three models are studied for this experiment: XLNet with the original segment recurrence, XLNet with utterance recurrence, and DialogXL. 

\begin{figure}[t]
	\centering
	\includegraphics[width=1\columnwidth]{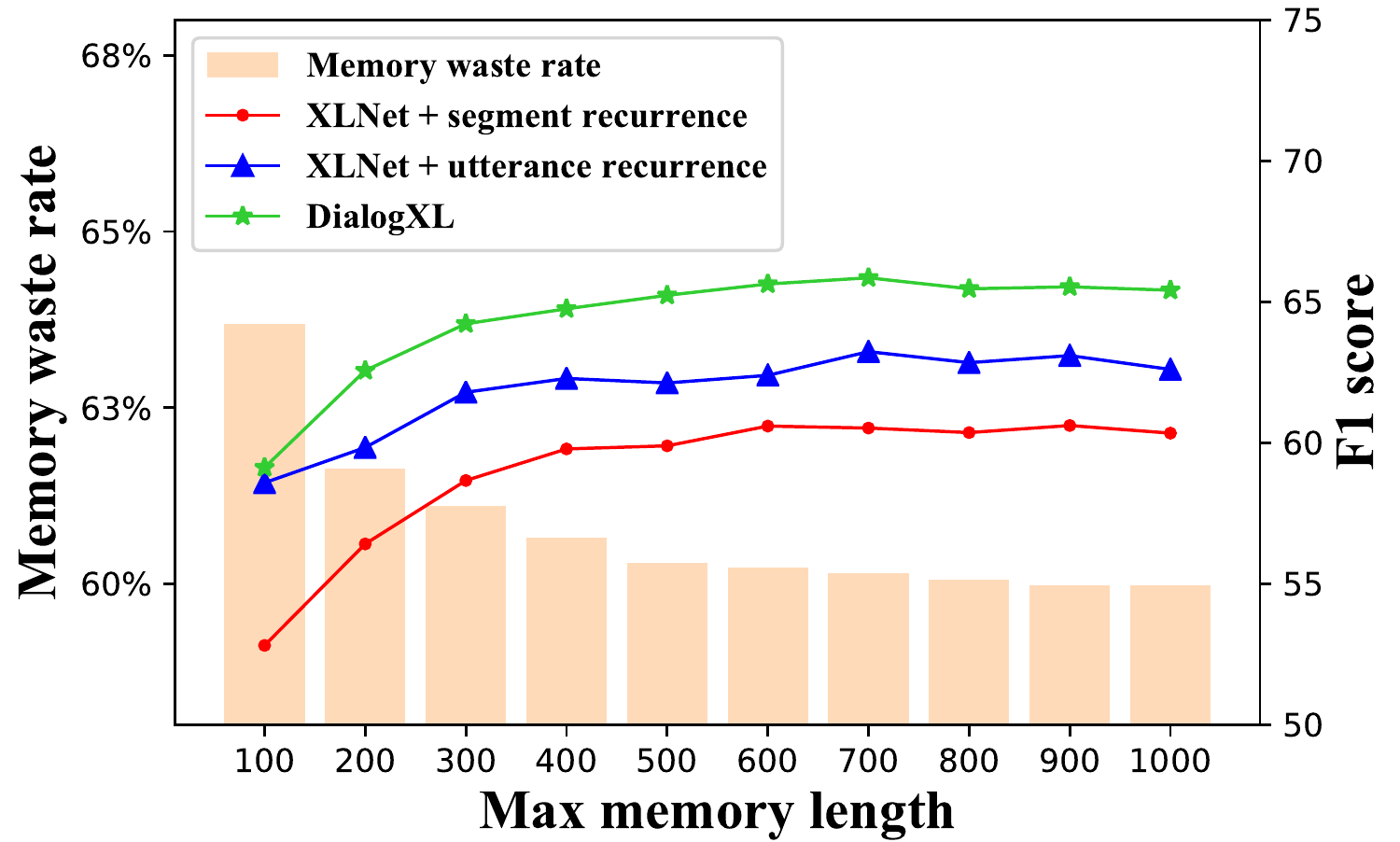} 
	\caption{The results of vanilla XLNet, XLNet with utterance recurrence, and our DialogXL by different maximum memory lengths on IEMOCAP. The memory waste rate of segment recurrence in XLNet  is provided for reference.}
	\label{fig:memory}
\end{figure}

The results are shown in Figure \ref{fig:memory}. We can note that segment recurrence always leads to a memory waste rate of over 60\% for each different memory length. The rate drops with the memory length increases, along with the growth of the three models. When the memory length exceeds 700, their performance generally stops improving any more, which indicates that increasing the maximum memory length only contributes to the test results within a certain range.

\subsection{Ablation Study}
In this ablation study, we analyze the impact of dialog-aware self-attention by removing each type of dialog-aware self-attention from DialogXL. The results on two representative datasets, IEMOCAP and MELD, are presented in Table \ref{tab:ablation}. 

We can observe that the performance of DialogXL drops on both IEMOCAP and MELD when any type of the self-attention is removed, suggesting that all these self-attentions contribute to the improvement of DialogXL. Nevertheless, their contributions can be distinguished. When speaker self-attention or listener self-attention is removed, considerable drops are observed. But when they are both removed, the drops are more obvious. This implies the importance of the inter-/intra-speaker dependency \cite{ghosal2019dialoguegcn}. 

Moreover, when local self-attention is removed, the F1 score drops the most on IEMOCAP, which contains long utterances (around 70) for each conversation. This indicates that the historical context near a query utterance is more important for this dataset. The drop on MELD is not as obvious as on IEMOCAP, because MELD has much shorter conversations (5 to 9 utterances per conversation). Finally, the removal of global self-attention leads to the least performance degradation. The reason could be twofold. First, global utterances are not as important as local utterances. Second, the speaker self-attention and listener self-attention already capture some useful information from distant utterances.

\begin{table}[t]
	\centering
	\resizebox{0.475\textwidth}{!}{
	\begin{tabular}{l|l|l}
		\hline
		\multirow{2}*{Method} & \multicolumn{2}{|c}{F1 score}\\ 
        \cline{2-3}
		&IEMOCAP & MELD\\
		\hline
		DialogXL & \textbf{65.94}& \textbf{62.41}\\
		- speaker self-attention&62.30 ($\downarrow$3.64)& 61.92 ($\downarrow$0.49)\\ 
		- listener self-attention& 62.87  ($\downarrow$3.07)&  62.03  ($\downarrow$0.38)\\ 
		- speaker\&listener self-attention& 61.71  ($\downarrow$4.23)& 61.70 ($\downarrow$\textbf{0.71})\\
		- local self-attention& 61.66  ($\downarrow$\textbf{4.28})& 61.72 ($\downarrow$0.69)\\
		- global self-attention& 63.34 ($\downarrow$2.60)& 62.15 ($\downarrow$0.26)\\
		\hline
	\end{tabular}
	}
	\caption{Results of ablation study on IEMOCAP and MELD.}
	\label{tab:ablation}
\end{table}

\begin{figure*}[t]
	\centering
	\includegraphics[width=1.9\columnwidth]{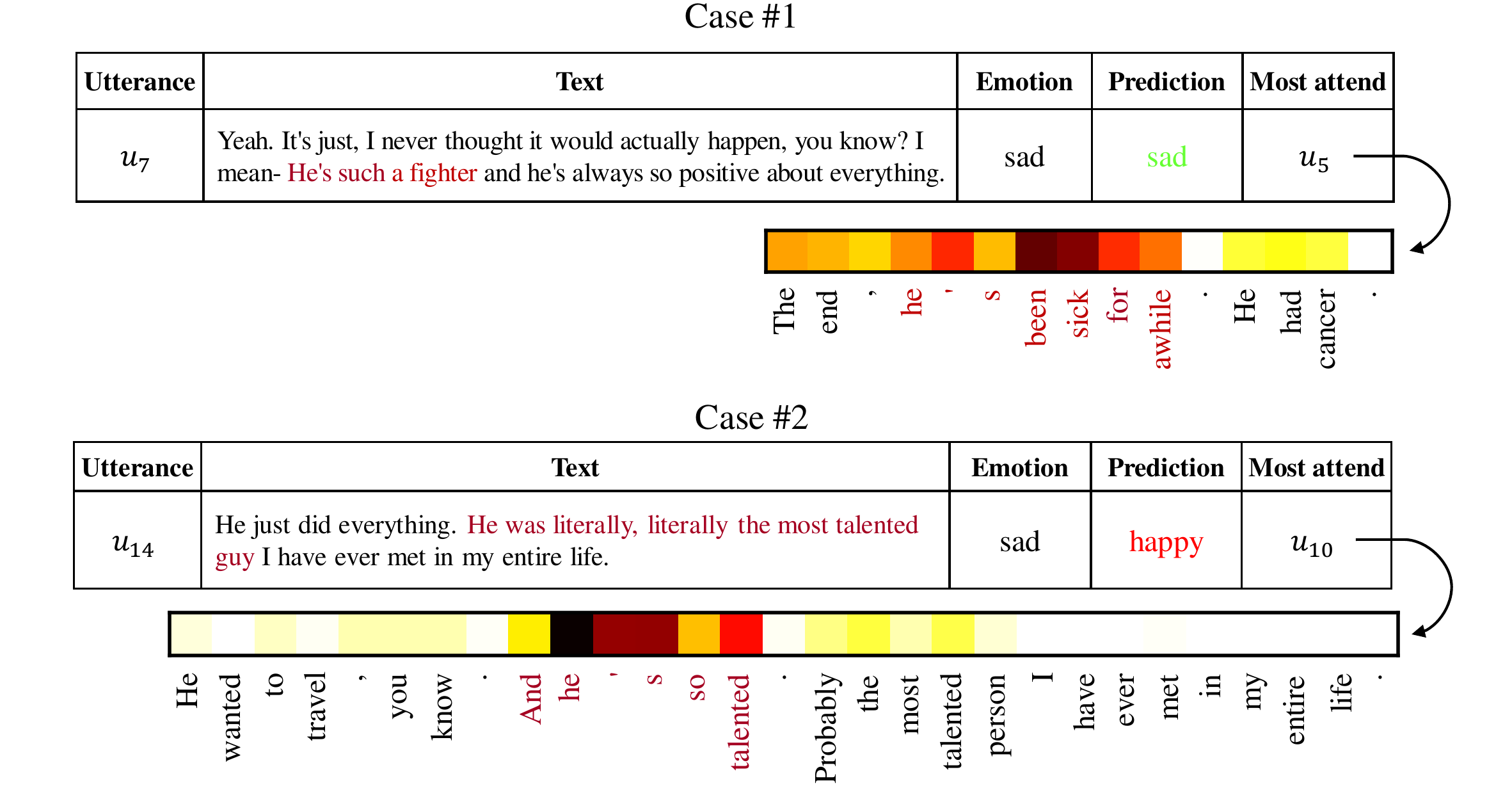} 
	\caption{Results of error analysis, where two query utterances are provided, along with the visualization of attention weights between the current utterance at the \text{[CLS]} token and the most attended historical utterance (selected according to the highest average attention weight among all the attention heads of the last layer). The darker colors mean larger attention weights.}
	\label{fig:visualization}
\end{figure*}

\subsection{Speaker Role Embedding}
Our speaker self-attention and listener self-attention model the speaker dependencies \cite{ghosal2019dialoguegcn} by directly letting the model know which part of the utterances should be attended to. Another way to let a pre-trained language model understand the speaker dependencies in dialog is speaker role embedding \cite{bao2020plato,ham2020end}, which maps each participant to a trainable embedding vector. Here, we make a simple comparison between the two approaches of embedding different parties on IEMOCAP and DailyDialog. To this end, we replace the speaker self-attention and listener self-attention of DialogXL with the speaker role embeddings, and refer to the resulting model as DialogXL-emb. The results of comparison are shown in Table \ref{tab:speker_role}. We can observe that our explicit speaker\&listencer self-attention is more effective than the speaker role embedding approach. As a result, the proposed attention mechanism can be potentially applied to other dialog tasks as well.

\begin{table}[h]
	\centering
		\resizebox{0.43\textwidth}{!}{
	\begin{tabular}{p{2.5cm}|p{2cm}<{\centering}|p{2cm}<{\centering}}
		\hline
		\multirow{2}*{Method} & \multicolumn{2}{c}{F1 score}\\ 
        \cline{2-3}
		&IEMOCAP & DailyDialog\\
		\hline
		DialogXL & \textbf{65.94}& \textbf{54.93}\\
		DialogXL-emb&63.31 & 54.06 \\ 
		\hline
	\end{tabular}
	}
	\caption{Results of comparison between direct speaker role embedding and our speaker\&listener self-attention approach on the IEMOCAP and DailyDialog datasets.}
	\label{tab:speker_role}
\end{table}

\subsection{Error Study}
Although our DialogXL has a novel framework and achieves a new state of the art, we still want to figure out its possible shortcomings to motivate the future research. Therefore, we carry out an error study on IEMOCAP. In short, we found that DialogXL's powerful capability of directly capturing word-level features in the historical context can be a double-edge sword. As illustrated in Figure \ref{fig:visualization}, the word-level attention mechanism based on semantic relevance can help make a good prediction (Case \#1), but it may also lead to a mistake by focusing too much on the semantic relevance between the query utterance and historical utterances (Case \#2). As a result, it seems to be necessary to combine with other mechanisms rather than merely relying on the popular attention to carry out the emotion recognition in dialogues.


Besides, we also observe from our bad cases that some of them are mentioned in previous works, such as emotional shifts (i.e., the emotion labels of two consecutive utterances from a same speaker are different)  \cite{hazarika2018conversational, majumder2019dialoguernn}. Roughly, our model commits mistakes for 45\% of these cases, which calls for further investigations. 

\section{Conclusion}
In this paper, we proposed an all-in-one XLNet model, namely DialogXL, for emotion recognition in conversation (ERC). To model the multi-turn multi-party conversational data, DialogXL contributes two improvements on the basis of XLNet and Transformer-XL. First, an enhanced memory was introduced to replace XLNet's vanilla memory to store historical contexts more effectively. Second, a dialog-aware self-attention mechanism was proposed to deal with the multi-turn multi-party data structures. Extensive experiments were conducted on four ERC benchmarks and the results show that the proposed model outperforms all the baselines on the datasets. The effectiveness of the two improvements is also confirmed by extensive analyses. Furthermore, we have the following three findings. First, the original segment recurrence mechanism stores more than 60\% paddings in memory, making it ineffective to encode the historical contexts for ERC. Second, the traditional speaker role embedding strategy is not as effective as our speaker\&listener self-attention, which could also be applied to other dialog tasks. Finally, an error analysis reveals that merely relying on the attention mechanism may mislead the model.

\section{Acknowledgments}
The paper was supported by the Fundamental Research Funds for the Central Universities (No.19lgpy220) and the Program for Guangdong Introducing Innovative and Entrepreneurial Teams (No.2017ZT07X355). This work was also supported by MindSpore.

\bibliography{ref.bib}

\begin{thebibliography}{26}
\providecommand{\natexlab}[1]{#1}
\providecommand{\url}[1]{\texttt{#1}}
\providecommand{\urlprefix}{URL }
\expandafter\ifx\csname urlstyle\endcsname\relax
  \providecommand{\doi}[1]{doi:\discretionary{}{}{}#1}\else
  \providecommand{\doi}{doi:\discretionary{}{}{}\begingroup
  \urlstyle{rm}\Url}\fi

\bibitem[{{Bao} et~al.(2020){Bao}, {He}, {Wang}, {Wu}, and
  {Wang}}]{bao2020plato}
{Bao}, S.; {He}, H.; {Wang}, F.; {Wu}, H.; and {Wang}, H. 2020.
\newblock PLATO: Pre-trained Dialogue Generation Model with Discrete Latent
  Variable.
\newblock In \emph{ACL 2020: 58th annual meeting of the Association for
  Computational Linguistics}, 85--96.

\bibitem[{Busso et~al.(2008)Busso, Bulut, Lee, Kazemzadeh, Mower, Kim, Chang,
  Lee, and Narayanan}]{busso2008iemocap}
Busso, C.; Bulut, M.; Lee, C.-C.; Kazemzadeh, A.; Mower, E.; Kim, S.; Chang,
  J.~N.; Lee, S.; and Narayanan, S.~S. 2008.
\newblock IEMOCAP: Interactive emotional dyadic motion capture database.
\newblock \emph{Language resources and evaluation} 42(4): 335.

\bibitem[{{Chen} et~al.(2018){Chen}, {Hsu}, {Kuo}, {Ting-Hao}, {Huang}, and
  {Ku}}]{chen2018emotionlines}
{Chen}, S.-Y.; {Hsu}, C.-C.; {Kuo}, C.-C.; {Ting-Hao}; {Huang}; and {Ku}, L.-W.
  2018.
\newblock EmotionLines: An Emotion Corpus of Multi-Party Conversations.
\newblock In \emph{11th International Conference on Language Resources and
  Evaluation, LREC 2018}, 1597--1601.

\bibitem[{{Conneau} et~al.(2020){Conneau}, {Khandelwal}, {Goyal}, {Chaudhary},
  {Wenzek}, {Guzmán}, {Grave}, {Ott}, {Zettlemoyer}, and
  {Stoyanov}}]{conneau2020unsupervised}
{Conneau}, A.; {Khandelwal}, K.; {Goyal}, N.; {Chaudhary}, V.; {Wenzek}, G.;
  {Guzmán}, F.; {Grave}, E.; {Ott}, M.; {Zettlemoyer}, L.; and {Stoyanov}, V.
  2020.
\newblock Unsupervised Cross-lingual Representation Learning at Scale.
\newblock In \emph{ACL 2020: 58th annual meeting of the Association for
  Computational Linguistics}, 8440--8451.

\bibitem[{{Dai} et~al.(2019){Dai}, {Yang}, {Yang}, {Carbonell}, {Le}, and
  {Salakhutdinov}}]{dai2019transformer}
{Dai}, Z.; {Yang}, Z.; {Yang}, Y.; {Carbonell}, J.; {Le}, Q.; and
  {Salakhutdinov}, R. 2019.
\newblock Transformer-XL: Attentive Language Models Beyond a Fixed-Length
  Context.
\newblock In \emph{ACL 2019 : The 57th Annual Meeting of the Association for
  Computational Linguistics}, 2978--2988.

\bibitem[{Devlin et~al.(2018)Devlin, Chang, Lee, and
  Toutanova}]{devlin2018bert}
Devlin, J.; Chang, M.-W.; Lee, K.; and Toutanova, K. 2018.
\newblock Bert: Pre-training of deep bidirectional transformers for language
  understanding.
\newblock \emph{arXiv preprint arXiv:1810.04805} .

\bibitem[{Ghosal et~al.(2019)Ghosal, Majumder, Poria, Chhaya, and
  Gelbukh}]{ghosal2019dialoguegcn}
Ghosal, D.; Majumder, N.; Poria, S.; Chhaya, N.; and Gelbukh, A. 2019.
\newblock DialogueGCN: A Graph Convolutional Neural Network for Emotion
  Recognition in Conversation.
\newblock In \emph{Proceedings of the 2019 Conference on Empirical Methods in
  Natural Language Processing and the 9th International Joint Conference on
  Natural Language Processing (EMNLP-IJCNLP)}, 154--164.

\bibitem[{{Ham} et~al.(2020){Ham}, {Lee}, {Jang}, and {Kim}}]{ham2020end}
{Ham}, D.; {Lee}, J.-G.; {Jang}, Y.; and {Kim}, K.-E. 2020.
\newblock End-to-End Neural Pipeline for Goal-Oriented Dialogue Systems using
  GPT-2.
\newblock In \emph{ACL 2020: 58th annual meeting of the Association for
  Computational Linguistics}, 583--592.

\bibitem[{Hazarika et~al.(2018)Hazarika, Poria, Mihalcea, Cambria, and
  Zimmermann}]{hazarika2018icon}
Hazarika, D.; Poria, S.; Mihalcea, R.; Cambria, E.; and Zimmermann, R. 2018.
\newblock Icon: Interactive conversational memory network for multimodal
  emotion detection.
\newblock In \emph{Proceedings of the 2018 Conference on Empirical Methods in
  Natural Language Processing}, 2594--2604.

\bibitem[{{Hazarika} et~al.(2018){Hazarika}, {Poria}, {Zadeh}, {Cambria},
  {Morency}, and {Zimmermann}}]{hazarika2018conversational}
{Hazarika}, D.; {Poria}, S.; {Zadeh}, A.; {Cambria}, E.; {Morency}, L.-P.; and
  {Zimmermann}, R. 2018.
\newblock Conversational Memory Network for Emotion Recognition in Dyadic
  Dialogue Videos.
\newblock In \emph{Proceedings of the 2018 Conference of the North American
  Chapter of the Association for Computational Linguistics: Human Language
  Technologies, Volume 1 (Long Papers)}, volume~1, 2122--2132.

\bibitem[{{Hazarika} et~al.(2019){Hazarika}, {Poria}, {Zimmermann}, and
  {Mihalcea}}]{hazarika2019emotion}
{Hazarika}, D.; {Poria}, S.; {Zimmermann}, R.; and {Mihalcea}, R. 2019.
\newblock Emotion Recognition in Conversations with Transfer Learning from
  Generative Conversation Modeling.
\newblock \emph{arXiv: Computation and Language} .

\bibitem[{Henderson et~al.(2019)Henderson, Casanueva, Mrk{\v{s}}i{\'c}, Su,
  Vuli{\'c} et~al.}]{henderson2019convert}
Henderson, M.; Casanueva, I.; Mrk{\v{s}}i{\'c}, N.; Su, P.-H.; Vuli{\'c}, I.;
  et~al. 2019.
\newblock ConveRT: Efficient and accurate conversational representations from
  transformers.
\newblock \emph{arXiv preprint arXiv:1911.03688} .

\bibitem[{Jiao et~al.(2019)Jiao, Yang, King, and Lyu}]{jiao2019higru}
Jiao, W.; Yang, H.; King, I.; and Lyu, M.~R. 2019.
\newblock HiGRU: Hierarchical Gated Recurrent Units for Utterance-Level Emotion
  Recognition.
\newblock In \emph{Proceedings of the 2019 Conference of the North American
  Chapter of the Association for Computational Linguistics: Human Language
  Technologies, Volume 1 (Long and Short Papers)}, 397--406.

\bibitem[{{Li} et~al.(2017){Li}, {Su}, {Shen}, {Li}, {Cao}, and
  {Niu}}]{li2017dailydialog}
{Li}, Y.; {Su}, H.; {Shen}, X.; {Li}, W.; {Cao}, Z.; and {Niu}, S. 2017.
\newblock DailyDialog: A Manually Labelled Multi-turn Dialogue Dataset.
\newblock In \emph{Proceedings of the Eighth International Joint Conference on
  Natural Language Processing (Volume 1: Long Papers)}, volume~1, 986--995.

\bibitem[{{Liu} et~al.(2019){Liu}, {Ott}, {Goyal}, {Du}, {Joshi}, {Chen},
  {Levy}, {Lewis}, {Zettlemoyer}, and {Stoyanov}}]{liu2019roberta}
{Liu}, Y.; {Ott}, M.; {Goyal}, N.; {Du}, J.; {Joshi}, M.; {Chen}, D.; {Levy},
  O.; {Lewis}, M.; {Zettlemoyer}, L.; and {Stoyanov}, V. 2019.
\newblock RoBERTa: A Robustly Optimized BERT Pretraining Approach.
\newblock \emph{arXiv preprint arXiv:1907.11692} .

\bibitem[{{Loshchilov} and {Hutter}(2018)}]{loshchilov2018fixing}
{Loshchilov}, I.; and {Hutter}, F. 2018.
\newblock Fixing Weight Decay Regularization in Adam .

\bibitem[{Madotto, Wu, and Fung(2018)}]{madotto2018mem2seq}
Madotto, A.; Wu, C.-S.; and Fung, P. 2018.
\newblock Mem2seq: Effectively incorporating knowledge bases into end-to-end
  task-oriented dialog systems.
\newblock \emph{arXiv preprint arXiv:1804.08217} .

\bibitem[{Majumder et~al.(2019)Majumder, Poria, Hazarika, Mihalcea, Gelbukh,
  and Cambria}]{majumder2019dialoguernn}
Majumder, N.; Poria, S.; Hazarika, D.; Mihalcea, R.; Gelbukh, A.; and Cambria,
  E. 2019.
\newblock Dialoguernn: An attentive rnn for emotion detection in conversations.
\newblock In \emph{Proceedings of the AAAI Conference on Artificial
  Intelligence}, volume~33, 6818--6825.

\bibitem[{{Poria} et~al.(2019){Poria}, {Hazarika}, {Majumder}, {Mihalcea},
  {Naik}, and {Cambria}}]{poria2019meld}
{Poria}, S.; {Hazarika}, D.; {Majumder}, N.; {Mihalcea}, R.; {Naik}, G.; and
  {Cambria}, E. 2019.
\newblock MELD: A Multimodal Multi-Party Dataset for Emotion Recognition in
  Conversations.
\newblock In \emph{ACL 2019 : The 57th Annual Meeting of the Association for
  Computational Linguistics}, 527--536.

\bibitem[{Poria et~al.(2019)Poria, Majumder, Mihalcea, and
  Hovy}]{poria2019emotion}
Poria, S.; Majumder, N.; Mihalcea, R.; and Hovy, E. 2019.
\newblock Emotion recognition in conversation: Research challenges, datasets,
  and recent advances.
\newblock \emph{IEEE Access} 7: 100943--100953.

\bibitem[{{Schuller} et~al.(2012){Schuller}, {Valster}, {Eyben}, {Cowie}, and
  {Pantic}}]{schuller2012avec}
{Schuller}, B.; {Valster}, M.; {Eyben}, F.; {Cowie}, R.; and {Pantic}, M. 2012.
\newblock AVEC 2012: the continuous audio/visual emotion challenge.
\newblock In \emph{Proceedings of the 14th ACM international conference on
  Multimodal interaction}, 449--456.

\bibitem[{{Vaswani} et~al.(2017){Vaswani}, {Shazeer}, {Parmar}, {Uszkoreit},
  {Jones}, {Gomez}, {Kaiser}, and {Polosukhin}}]{vaswani2017attention}
{Vaswani}, A.; {Shazeer}, N.; {Parmar}, N.; {Uszkoreit}, J.; {Jones}, L.;
  {Gomez}, A.~N.; {Kaiser}, L.; and {Polosukhin}, I. 2017.
\newblock Attention is All You Need.
\newblock In \emph{Proceedings of the 31st International Conference on Neural
  Information Processing Systems}, 5998--6008.

\bibitem[{Yang et~al.(2019)Yang, Dai, Yang, Carbonell, Salakhutdinov, and
  Le}]{yang2019xlnet}
Yang, Z.; Dai, Z.; Yang, Y.; Carbonell, J.; Salakhutdinov, R.~R.; and Le, Q.~V.
  2019.
\newblock Xlnet: Generalized autoregressive pretraining for language
  understanding.
\newblock In \emph{Advances in neural information processing systems},
  5753--5763.

\bibitem[{{Zahiri} and {Choi}(2017)}]{zahiri2017emotion}
{Zahiri}, S.~M.; and {Choi}, J.~D. 2017.
\newblock Emotion Detection on TV Show Transcripts with Sequence-based
  Convolutional Neural Networks.
\newblock In \emph{AAAI Workshops}, 44--52.

\bibitem[{Zhong, Wang, and Miao(2019)}]{zhong2019knowledge}
Zhong, P.; Wang, D.; and Miao, C. 2019.
\newblock Knowledge-Enriched Transformer for Emotion Detection in Textual
  Conversations.
\newblock In \emph{Proceedings of the 2019 Conference on Empirical Methods in
  Natural Language Processing and the 9th International Joint Conference on
  Natural Language Processing (EMNLP-IJCNLP)}, 165--176.

\bibitem[{Zhou et~al.(2017)Zhou, Huang, Zhang, Zhu, and
  Liu}]{zhou2017emotional}
Zhou, H.; Huang, M.; Zhang, T.; Zhu, X.; and Liu, B. 2017.
\newblock Emotional chatting machine: Emotional conversation generation with
  internal and external memory.
\newblock \emph{arXiv preprint arXiv:1704.01074} .

\end{thebibliography}
\end{document}